\documentclass[times,11pt]{article}
\usepackage{cyc}
\usepackage{epsf}

\makeatletter
\def\section{\@startsection {section}{1}{\z@}{18pt plus
    2pt minus 2pt}{1.5ex plus 0.3ex minus .2ex}{\Large\bf\raggedright}}
\def\subsection{\@startsection{subsection}{2}{\z@}{12pt plus
    2pt minus 2pt}{0.8ex plus .2ex}{\large\bf\raggedright}}
\makeatother
\begin{document}

\centerline{\LARGE\bf Semantic Parsing based on Verbal Subcategorization} 
{~}\\[10pt]
{\large\it  Jordi Atserias} \\
{\large\it  Irene Castell\'on} \\ 
{\large\it  Montse Civit} \\
{\large\it  German Rigau} \\[10pt]
\noindent

The aim of this work is to explore new methodologies on Semantic
Parsing for unrestricted texts. 
Our approach follows the current trends in Information Extraction (IE)
and is based on the application of a verbal subcategorization lexicon
(LEXPIR) by means of complex pattern recognition techniques. LEXPIR is
framed on the theoretical model of the verbal subcategorization
developed in the Pirapides project.
  
\section{{~}{~}I{\large ntroduction}}

Most of the different tasks included in Natural Language Processing
(such as Information Retrieval, Information Extraction, Information
Filtering, Natural Language Interfaces and Story Understanding) apply
different levels of Natural Language Understanding.
For instance, in the case of Information Extraction, the Natural
Language Understanding component plays a crucial role. This is due to
the fact that most of the information to be extracted can only be
identified by recognizing the conceptual roles. This area has been
greatly promoted by the Message Understanding Conferences (MUC's)
organized by TIPSTER.

Such conferences have shown the tendency of the Information Extraction
Systems to be more domain \cite{Wilks+'99} and language independent
\cite{MLASIE} \cite{Kilgariff'97}, making Information Extraction stand closer to
Natural Language Understanding. Currently, other related areas (such
as Story Understanding \cite{Riloff'99}) have begun to adapt the
recent improvements done in Information Extraction.

An important step in any process that implies Natural Language
Understanding is Semantic Interpretation.  Semantic Interpretation can
be defined as the process of obtaining a suitable meaning
representation for a text.
The input of the Semantic Interpreter can vary largely, going from raw
text to full parsing trees. Likewise, the output of the Semantic
Interpreter can also vary considerably (logical formulae, case-frames,
SQL), mostly influenced by the type of application. In relation to
this, two important sub-tasks can be distinguished within Semantic
Interpretation: Word Sense Disambiguation (WSD) and Semantic Parsing,
being the latter the interest of the current work. Further, an
essential part of the Semantic Parsing involves the production of a
case-role analysis in which the semantic roles of the entities, such
as \emph{starter} or \emph{instrument}, are identified
\cite{Brill+Mooney'97}.

The work here presented focuses on this problem, in particular on the
issue of obtaining the verbal argument structure of the sentence. 
Our proposal for
obtaining the representation of the meaning components (roles) of the
verb is based on the application of the linguistic theory of the
verbal subcategorization developed inside the Pirapides Project
\cite{Fernandez+'99}, and is performed by means of complex pattern
recognition techniques.

Pirapides is a project centered on the study of the English, Spanish and
Catalan verbal predicates. Pirapides has several goals: On the one
hand, from a theoretical point of view,  a deep study is being carried
out of the units that the model of a verbal entry has
produced. This syntactic component focuses on the representation of the interaction
between the syntactic and semantic components. 

On the other hand, from
an application-oriented point of view, a lexicon (LEXPIR) is being
developed, based on this theoretical model, which will be used to
analyze the corpus.
%LEXPIR includes different kinds of information associated to each predicate.

Following this brief introduction, Section 2 presents the linguistic
model and Section 3 the computational model. Then, Section 4 describes
the experiments carried out and the results obtained. Finally, Section
5 draws some conclusions and presents further work.
%-----------------------------------------------------------------------------
\section{{~}{~}L{exical Model}}

The syntactic analysis using Context Free Grammars (CFG) for non
domain-specific Spanish corpora has several limitations: it is
basically impossible to carry out an analysis at a sentence level
including syntactic functions. This is mainly due to the optionality
of some constituents (such as the subject), and also because of the
free order of the constituents.

Further, phrase analysis is not enough in order to obtain a suitable
interpretation of the sentence. Thus, it becomes necessary to explore
new tools to go beyond the phrase level.

Bearing this goal in mind, a hierarchical verbal lexicon for Spanish (LEXPIR) is being developed.
In this lexicon, verbs are
grouped hierarchaly based on their meaning components as well as their diathetic
alternations \cite{Fernandez+'96} \cite{Fernandez+'99},
\cite{Morante+'98}. Moreover, each group is subclassified according to
the number of components which can be explicitly realized. In
addition, LEXPIR includes, for each verb sense, information about the
number of arguments, their syntactic realization, the prepositions
they can take, their semantic component, their agreement and their
optionality.

The information is propagated within the hierarchy in a top-down
manner, that is, each verb inherits the elements from its group and
each group from its class. However, the inherited information can be
overwritten by the information already associated to the specific verb
entry (default monotonic inheritance).
%
%--------- Figura ------------  
%
\begin{figure*}[htb]
\begin{center}
\epsfxsize=\linewidth
\hfil\epsfbox{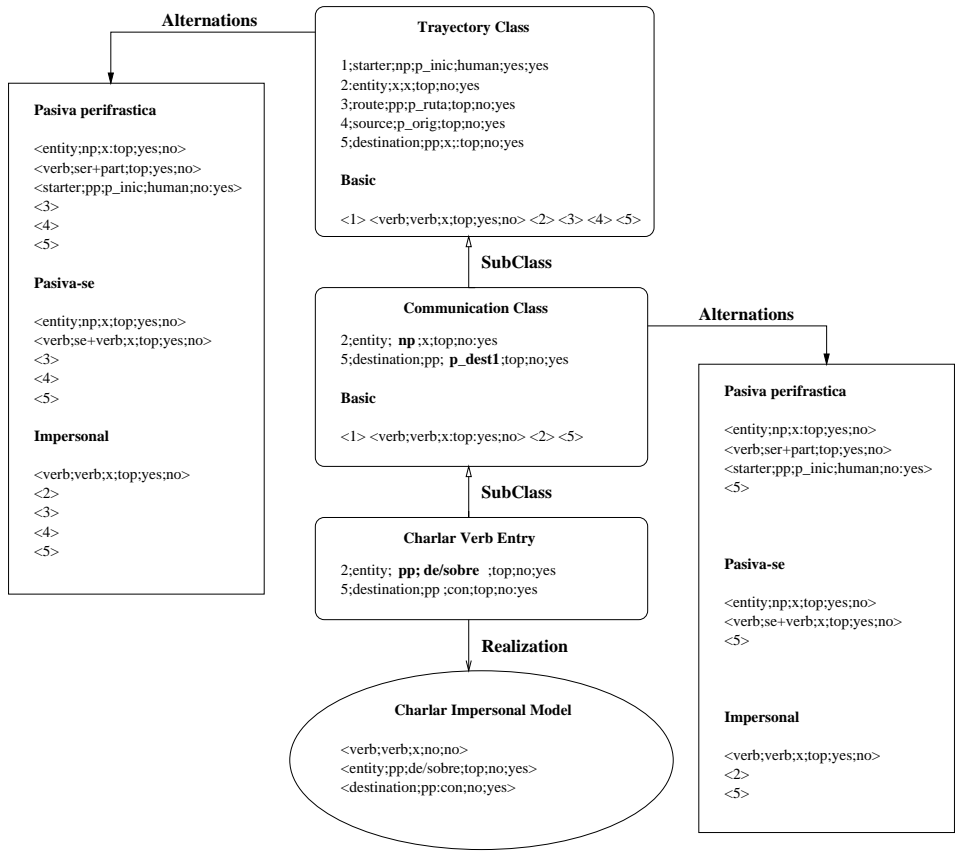}\hfil
\caption{ \label{classes} Class Hierarchy}
\end{center}
\end{figure*}
%----------------- TAULA --------------
\begin{table*}[hbt] \centering
\begin{tabular}{|l|l|l|l|l|l|l|}
\hline
\textbf{N$^o$ Id} & \textbf{Syntax} & \textbf{Prep.} & \textbf{Component} & \textbf{Semantics} & \textbf{Agree.} &
\textbf{Opt.} \\ \hline
1 & NP & p\_inic & starter & Human & yes & yes \\ \hline
2 & x  & x 	 & entity  & Top   & no  & yes \\ \hline
3 & PP & p\_ruta & route   & Top   & no  & yes \\ \hline 
4 & PP & p\_orig & source  & Top    & no &  yes \\ \hline
5 & PP & x	 & destination & Top & no & yes \\ \hline
\end{tabular}
\caption{Basic Model for trajectory verbs}
\label{example}
\end{table*}

As shown in Figure \ref{classes}, the trajectory class has five
components: \emph{starter}(1), \emph{entity}(2) and the \emph{trajectory},
which is the component which defines the class. The trajectory can be
further divided into three components: \emph{route}(3), \emph{source}(4) and
\emph{destination}(5). Each one of these components has a basic phrase
structure, a set of prepositions introducing them and a particular
semantics. Moreover, one of the components must be in agreement with
the verb.
\pagebreak

There are four subclasses included in the trajectory class:
\emph{non-autonomous movement}, \emph{autonomous movement},
\emph{communication} and \emph{transfer} (although the last one has
not been formalized yet).

The non-autonomous movement subclass is characterised by the fact that
it explicitly realizes the five components: ``Alguien (1) desplaza
algo (2) por un lugar (3) desde un punto (4) a otro (5)'' (somebody
moves something through a place from one point to another).

The autonomous movement presents a co-indexation between the
components \emph{starter} and \emph{entity}.
E.g.: ``Alguien (1,2) va por un lugar (3) desde un punto (4) a otro
(5)'' (somebody goes through a place from one point to another).  As
it can be seen in the example, (1) is at the same time the
\emph{starter} of the event \emph{go} and the \emph{entity} that is moved.

Finally, in the communication class there are only three components
which are explicit: \emph{starter}, which is at the same time the
\emph{source}, \emph{entity} and \emph{destination}). E.g.: ``Alguien
(1,4) dice algo (2) a alguien (5)'' (somebody says something to
somebody).

Regarding prepositions, those which can appear in the destination
component (5) are specified in the subclasses and can be divided into
two groups: ``p\_dest1'' which includes ``a/para'' (to) and
``p\_dest2'' which includes the rest of preposition for \emph{destination}.

Moreover, specific verb forms can impose their own restrictions. For instance,
``charlar'' (to chat) is a verb which, in contradiction with the rest
of the verbs in the communication class, does not accept an NP in the
entity component and the PP must have the preposition ``de/sobre''
(about). Furthermore, it cannot take the prepositions ``a/para'' to
express destination and uses the preposition ``con'' instead.  E.g.:
``Alguien (1) charla de algo (2) con alguien (5)'' (somebody chats
about something with somebody).

Finally, in order to obtain the alternation schemes for a verb, the
information of the verb is composed with the alternations of the
class. The different elements that appear on a model are explained
below for a specific case: the basic model for the trajectory verbs
(see Table \ref{example}).
\noindent
\begin{itemize}
\item \emph{Id Number}: Numeric value that identifies the meaning component.
\item \emph{Syntax}: Syntactic realization of the semantic
component. For the second component this information is unspecified (x)
as the syntactic realization depends on the subclass. Moreover, this
element, which is usually the Direct Object, has other restrictions:
if its semantics indicates that it is [+human/animate] it should be a
PP, while if it is [-human/animate] it has to be realized as an NP.
\item \emph{Preposition}: List of prepositions which have been established
according to their meanings and occurrences.
\item \emph{Component}: Meaning component determined by the class.
\item \emph{Semantics}: Semantics of the component; this is a feature
specific of the argument.
\item \emph{Agreement}: Person and number agreement with the verb.
\item \emph{Optionality}: This indicates which elements are optional
inside the sentence.
\end{itemize}   
%----------------------------
Treating the optionality of the meaning components within
the model itself allows us to reduce the number of possible
alternations which have been established at a theoretical level
(Pirapides takes the underspecification of a component as an
alternation). Only that information which is different to the one
association to the class is actually marked.
For instance, in the \textbf{Pasiva perifr\'astica} model associated to the
communication class (see Figure \ref{classes}), the entity element  
(defined as \{entity;\textbf{NP};x;Top:\textbf{yes};no\}), 
has to be realized as an NP and also has to agree with the verb, which
is not the usual case in the communication class.
%-----------------------------------------------------------------------------
\section{{~}{~}C{omputational Model}}

LEXPIR allows the construction of patterns for all the possible
syntactico-semantic alternations of a verb. However, our goal is to
identify the meaning components of these patterns among the components
of the partial parsing tree of a sentence. Simultaneous to the
selection of the most similar verbal scheme for the sentence, the
meaning components are also obtained.

Due to the richness of the language (adjuncts, free order, etc.) there
is a need to apply robust pattern recognition techniques which allow to
change the position of some elements, the absence of certain elements
or the presence of new elements.  The following subsections focus on
the definition of the technique used for recognizing these complex
structures within a sentence.
%--------------------------------------------------------------------------
\subsection{{~}{~}A{pproximate pattern matching}}

The use of full parsing trees \cite{Atserias+'99} implies previous
decisions on the relationship between elements (e.g. PP-attachments). 
A mis-identified syntactic component, or whose limits have not been
correctly set, makes difficult not only the recognition of the
meaning components but also the recognition of the model itself.

To avoid this problem it was decided to use a syntactic analysis based on
syntactic unambiguous groups: chunks \cite{Abney'91}. This turns the
problem of comparing phrase structure trees into a problem of aligning
phrase group sequences. 
%----------------------------------------------------------------------------
\subsection{{~}{~}S{imilarity measures}}

Our similarity measure is defined in terms of the minimum cost
sequence of editing operations that transforms one structure into the
other.  The main differences with previous works on approximate
pattern matching based on editing operations \cite{Tsong-li+'94},
\cite{Shasha+'94} is that the elements in our sequences are Feature
Structures (FS). So the relabelling operation is performed on the
attributes.

As a consequence, the following editing operations were defined:
\begin{itemize}
\item \emph{Delete}: Deletes an element of the sequence.
\item \emph{Insert}: Inserts a new element in the sequence. 
\item \emph{Move}: Changes the order of an element in the sequence
(e.g.: ``[We] [went] [to Barcelona] [by plane]'' and ``[by plane] [We] [went] [to Barcelona]'').
\item \emph{Relabel}: Changes the value of the feature (attribute) of
an element in the sequence.
\end{itemize}

The cost of a sequence of operations is the addition of the cost of
each operation. In order to avoid having to choose the smallest model,
a correction factor inversely proportional to the number of nodes is
added to the similarity measure. It should be pointed out that the
number of Relabel and Delete operations gives a measure of the
goodness of the matching while the number of Insert operations
measures how much information from the sentence is not captured by the
pattern.
%-----------------------------------------------------------------------------
\section{{~}{~}E{xperiments}}

The experiments here presented aim to prove not only the feasibility
of the linguistic and computational models but also the possibility to
apply the system for improving and developing the verbal
subcategorization lexicon (LEXPIR).

In order to carry out the experiments a preliminary version of LEXPIR
was manually built, which contained 61 verbs belonging to the
trajectory class. Then, 170 sentences taken from an Spanish newspaper
were labelled by hand with the verbal models and the meaning
components. It should be also mentioned that only three sentences
present more than one model.
 
\subsection{{~}{~}P{rocessing the corpus}}

The corpus was pre-processed automatically to obtain a parsed tree for
each sentence. Firstly, the corpus was morphologically analized
\emph{MACO} \cite{Carmona+'98} and disambiguated \emph{Relax}
\cite{Padro'98}).  Secondly, the Spanish Wordnet \cite{Rodriguez+'98}
was used to semantically annotate the corpus with the 79 semantic
labels defined in the preliminary version of EuroWordnet Top
Ontology. Then, in order to obtain a partial parsing a context free
parser based on charts \emph{TACAT} and a wide coverage grammar of Spanish\footnote{Several rules were added to the grammar so as to deal with
noun complements.} \cite{Castellon+'98} were used to obtain the
partial parsing trees (see Figure \ref{example}).
Finally, those parsed tree were used by our system
to produce a case-role representation of the meaning components.
%------------- FIGURA 1-----------------
%\onecolumn
\begin{figure}[htb]
\begin{center}
\hfil\epsfbox{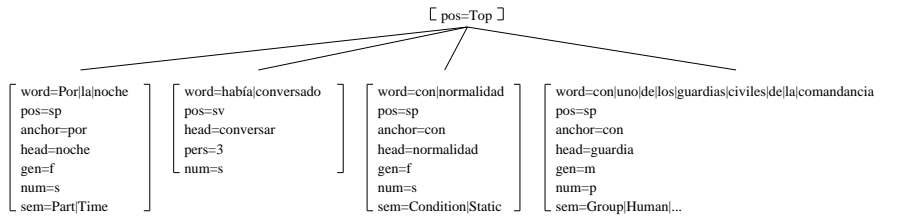}\hfil
\caption{ \label{arbre} Partial Parse Tree of ``At night (He/She) had
talked with one of the policemen from the commander's headquarters''}
\end{center}
\end{figure}
%\twocolumn
\begin{figure}[htb]
\begin{tabular}{|l|l|}
\hline 
\textbf{Meaning Comp.} & \textbf{Lexical Group} \\ \hline
\textbf{Event} & hab\'{\i}a conversado \\ \hline
\textbf{Destination} & con uno de los guardias \\ 
                    & civiles de la comandancia \\ \hline
\end{tabular}
\caption{\label{result} Meaning components obtained with the Basic Model}
\end{figure}
For instance, Figure \ref{result} shows the feature structure of the
meaning components obtained from the parsed tree shown in Figure \ref{arbre}.

% ----------------- Figura 2 ----------------------
\subsection{Evaluation \& Results}

The evaluation of a system which performs semantic
interpretation is a difficult task.  One of the contributions
of the MUC's has been to establish a set of evaluation metrics and a common
frame for the evaluation of Information Extraction Systems.  The MUC
evaluation methodology is based on a pre-alignment of the entities
from the solution and response.

However, in order to evaluate the results of our system, the existence
of two main differences has to be taken into account: multiple
instantiation and entity fragmentation.

\begin{itemize}
\item \emph{Multiple instantiation of the same model (entity)}: The
generation of different instantiations of the same entity is unusual
in Information Extraction, while our system does so.  For instance,
for the sentence ``[Pedro] [habl\'o] [con normalidad] [con Andr\'es]''
(Pedr\'o talked normally with Andr\'es), two solutions of the basic
model are obtained, one filling the role \emph{entity} with
\emph{normalidad} and the other with \emph{Andr\'es}.
\item \emph{Entity Fragmentation}: On the other hand, IESs do not
always recognize an entity as a whole, so that they generated several
entities corresponding to the different fragments. In our system this
could not happen as only a model per sentence is considered.
\end{itemize}

Assuming the existence of only one correct instantiation of a model
per sentence, our pre-alignment method consists in comparing all the
answers of the same model with the corresponding solution. As in
MUC-7, a role is correct if, and only if, both values are equal as strings.

Table \ref{resrole} shows the results in the identification of the
meaning components corresponding to verb arguments and applying the
MUC-7 evaluation metrics (see Table \ref{mucscorers}). Further, Table
\ref{resmodel} shows the results obtained on the identification of the
verb model . It should be mentioned that due to errors in the
pre-processing of the corpus, the system was unable to identify any
model for 5 of the 170 sentences.
%------------------ TABLES -------------------------
\newline
\noindent
\begin{table}[htb] \centering
\begin{tabular}{|c|c|c|c|}
\hline
\emph{COR} & \emph{INC} & \emph{PRE} & \emph{REC} \\ \hline
158 & 10 & 0.94 & 0.91 \\ \hline
\end{tabular}
\caption{Model Identification Results}
\label{resmodel}
\end{table}
\begin{table}[htb] \centering
\begin{tabular}{|c|c|c|c|c|}
\hline
\textbf{COR} & \textbf{INC} & \textbf{MIS} & \textbf{SPU} & \textbf{POS} \\ \hline
210 & 37 & 89 & 52 & 336 \\ \hline
\textbf{ACT} & \textbf{PRE} & \textbf{REC} & \textbf{UND} & \textbf{OBV} \\ \hline
299 & 0.7 & 0.6 & 0.26 & 0.17 \\ \hline
\textbf{SUB} & \textbf{ERR} & \textbf{P\&R} & \textbf{2P\&R} & \textbf{P\&R} \\ \hline
0.15 & 0.46 & 0.66 & 0.64 & 0.69 \\ \hline   
\end{tabular}
\caption{Meaning Components Results}
\label{resrole}
\end{table}
\pagebreak
%-----------------------------------------------------------------------------
%\section{MUC-7 measures}

\begin{table*}[htb] \centering
\begin{tabular}{|c|l|c|}
\hline
\small
\textbf{COR} & Number correct & \\ \hline
\textbf{INC} & Number incorrect & \\ \hline
\textbf{MIS} & Number missing &\\ \hline
\textbf{SPU} & Number spurious & \\ \hline
%\end{tabular}
%\begin{tabular}{|c|l|c|}
%\hline
\textbf{POS} & Number possible (elements in the solution) & \tiny
$COR+INC+MIS$ \small \\ \hline
\textbf{ACT} & Number actual (elements in the response) & \tiny
$COR+INC+SPU$ \small \\ \hline
%\end{tabular}    
%\begin{tabular}{|c|l|c|}
\hline 
\textbf{REC} & Recall & \tiny $\frac{COR}{POS}$ \small \\ \hline
\textbf{PRE} & Precision & \tiny$\frac{COR}{ACT}$ \small \\ \hline
\textbf{UND} & Undergeneration & \tiny $\frac{MIS}{POS}$ \small \\ \hline
\textbf{OVG} & Overgeneration & \tiny $\frac{SPU}{ACT}$ \small \\ \hline
\textbf{SUB} & Substitution & \tiny $\frac{INC}{COR+INC}$ \small \\ \hline
\textbf{ERR} & Error per response fill & \tiny $\frac{INC+SPU+MIS}{COR+INC+SPU+MIS}$ \small \\ \hline
\textbf{F-MESURES} & Weighted combination of REC \& PRE & \tiny
$\frac{(B^2 +1.0) \times P \times R}{(B^2 \times P)+R}$ \small \\ \hline
\end{tabular}
\normalsize
\caption{MUC-7 Evaluation Metrics}
\label{mucscorers}
\end{table*}
%-----------------------------------------------------------------------------
\section{{~}{~}C{onclusions \& Further Work}}

This paper has presented a semantic parsing approach for non
domain-specific texts. Our approach is based on the application of a
verbal subcategorization lexicon (LEXPIR) developed in the Pirapides
project.

The results of the experiment are very promising. Even though they
have been carried out using a limited corpus and lexicon, they have
proved the feasibility of the linguistic and computational models.

As further work it is planned to cover linguistic phenomena other than
the verbal subcategorization and to expand our system to deal with the
combination of multiple models beyond the usual cascade approach.  To
design a more general framework, it has also been planned to formalize
the role identification and model combination processes as a
Consistency Labelling Problem \cite{Pelillo+Refice'94,Padro'98} in which different
nominal and verbal models can compete for their case-role assignments.

%\pagebreak
%------------------------------------------------------------------------------
\section{{~}{~}A{knowledgments}}
\small This research has been partially funded by the Spanish Research
Department (Spontaneous-Speech
Dialogue System for Limited Domains TIC98-423-C06)
\normalsize
%-----------------------------------------------------------------------
%\pagebreak
\bibliographystyle{acl}
\bibliography{jab}

\begin{thebibliography}{}

\bibitem[\protect\citename{Abney}1991]{Abney'91}
Steven Abney, 1991.
\newblock {\em Parsing by chunks}.
\newblock Kluwer Academic Publishers.

\bibitem[\protect\citename{Atserias \bgroup et al.\egroup }1999]{Atserias+'99}
J.~Atserias, I.~Castell\'on, M.~Civit, and G.~Rigau.
\newblock 1999.
\newblock Using diathesis for semantic parsing.
\newblock In {\em Proceedings of Venecia per il Trattamento automatico delle
  lingue (VEXTAL)}, pages 385--392, Venecia, Italy.

\bibitem[\protect\citename{Brill and Mooney}1997]{Brill+Mooney'97}
Eric Brill and Raymond~J. Mooney.
\newblock 1997.
\newblock {An Overview of Empirical Natural Language Processing}.
\newblock {\em {Artificial Intelligence Magazine}}, 18(14):13--24, Winter.
\newblock Special Issue on Empirical Natural Language Processing.

\bibitem[\protect\citename{Carmona \bgroup et al.\egroup }1998]{Carmona+'98}
J.~Carmona, S.~Cervell, L.~M\'arquez, M.A. Mart\'{\i}, L.~Padr\'o, R.~Placer,
  H.~Rodr\'{\i}eguez, M.~Taul\'e, and J.~Turmo.
\newblock 1998.
\newblock { An Environment for Morphosyntactic Processing of Unrestricted
  Spanish Text}.
\newblock In {\em Proceedings of the First International Conference on Language
  Resources and Evaluation (LREC'98)}, Granada, Spain.

\bibitem[\protect\citename{Castell\'on \bgroup et al.\egroup
  }1998]{Castellon+'98}
Irene Castell\'on, Montse Civit, and Jordi Atserias.
\newblock 1998.
\newblock Syntactic parsing of spanish unrestricted text.
\newblock In {\em Proceedings of the 1th Conference on Language Resources and
  Evaluation (LREC'98)}, Granada. Spain.

\bibitem[\protect\citename{Fern\'andez and Mart\'{\i}}1996]{Fernandez+'96}
A.~Fern\'andez and M.~A. Mart\'{\i}.
\newblock 1996.
\newblock Classification of psycological verbs.
\newblock {\em SEPLN}, (20).

\bibitem[\protect\citename{Fern\'andez \bgroup et al.\egroup
  }1999]{Fernandez+'99}
A.~Fern\'andez, M.~A. Mart\'{\i}, G.~V\'azquez, and I.~Castell\'on.
\newblock 1999.
\newblock Establising semantic oppositions for typification of predicates.
\newblock {\em Language Design}, (2).

\bibitem[\protect\citename{Humphreys \bgroup et al.\egroup }1998]{MLASIE}
K.~Humphreys, R.~Gaizauskas, S.~Azzam, C.~Huyck, B.~Mitchell, H.~Cunningham,
  and Y.~Wilks.
\newblock 1998.
\newblock University of sheffield: Description of the lasie-ii system as used
  for muc-7.
\newblock In {\em MUC-7}.

\bibitem[\protect\citename{Kilgariff}1997]{Kilgariff'97}
Adam Kilgariff.
\newblock 1997.
\newblock Foreground and background lexicons and word sense disambiguation for
  information extraction.
\newblock In {\em Proceedings of the Workshop on Lexicon Driven Information
  Extraction}, Frascati, Italy.

\bibitem[\protect\citename{Morante \bgroup et al.\egroup }1998]{Morante+'98}
R.~Morante, Irene Castell\'on, and Gloria V\'azquez.
\newblock 1998.
\newblock Los verbos de trayectoria.
\newblock In {\em Proceedings of the conference of the SEPLN}.

\bibitem[\protect\citename{Padr\'o}1998]{Padro'98}
Lluis Padr\'o.
\newblock 1998.
\newblock {\em WSD Relaxation Labelling}.
\newblock {Ph.D.} thesis, Universitat Politectnica de Catalunya.

\bibitem[\protect\citename{Pelillo and Refice}1994]{Pelillo+Refice'94}
M.~Pelillo and M.~Refice.
\newblock 1994.
\newblock Learning compatibility coefficients for relaxation labelling
  processes.
\newblock {\em IEEE Transactions on Pattern Analysis and Machine Intelligence},
  16(9).

\bibitem[\protect\citename{Riloff}1999]{Riloff'99}
Ellen Riloff, 1999.
\newblock {\em Information Extraction as a Stepping Stone toward Story
  Understanding}.
\newblock MIT press, Montreal, Canada.

\bibitem[\protect\citename{Rodr\'{\i}guez \bgroup et al.\egroup
  }1998]{Rodriguez+'98}
Horacio Rodr\'{\i}guez, Salvador Climent, Peek Vossen, L.~Blocsma, Wim Peters,
  A.~Alonge, F.~Bertagna, and A.~Rovertini.
\newblock 1998.
\newblock The top-down strategy for building euwn: Vocabulary coverage, base
  concepts and top ontology.
\newblock {\em Computers and the Humanities}, 32(2-3).

\bibitem[\protect\citename{Shasha \bgroup et al.\egroup }1994]{Shasha+'94}
D.~Shasha, J.~Tson-Li Wang, K.~Zhang, and Y.~Shih.
\newblock 1994.
\newblock Exact and approximate algorithms for unorder tree matching.
\newblock {\em IEEE transactions on System\, Man and Cybernetics},
  28(5):668--678, April.

\bibitem[\protect\citename{Tsong-li \bgroup et al.\egroup }1994]{Tsong-li+'94}
J.~Tsong-li, K.~Zhang, K.~Jeong, and D.~Shasha.
\newblock 1994.
\newblock A system for approximate tree matching.
\newblock {\em IEEE transactions on Knowledge and Data Engineering}, 6(4).

\bibitem[\protect\citename{Wilks and Catizone}1999]{Wilks+'99}
Yorick Wilks and Roberta Catizone, 1999.
\newblock {\em Can We Make Information Extraction More Adaptative}, pages
  1--16.
\newblock Lecture Notes in artificial Intelligence. Springer-Verlang.
\newblock Subseries of Lectures Notes in Computer Science.

\end{thebibliography}
{~}\\

\noindent
\small {\bf\it Jordi Atserias} is a doctoral student at the Software
Department, Universitat Politecnica de Catalunya.
{\it batalla@lsi.upc.es}; URL: {\it http://www.lsi.upc.es/$\sim$batalla}

\noindent
\small {\bf\it Irene  Castell\'on} is a researcher and professor at
the Universitat de Barcelona. 
{\it castel@lingua.fil.ub.es}; URL: {\it http://www.ub.es/ling/labcat.htm}

\noindent
\small {\bf\it Montse Civit} is a doctoral student at the Software
Department, Universitat de Barcelona.
{\it civit@lsi.upc.es}; URL: {\it http://www.lsi.upc.es/$\sim$civit}

\noindent
\small {\bf\it  German Rigau}is a researcher and professor at the Software
Department, Universitat Politecnica de Catalunya.
{\it g.rigau@lsi.upc.es}; URL: {\it http://www.lsi.upc.es/$\sim$rigau}
\end{document}